# Practical model-based diagnosis with qualitative possibilistic uncertainty

**Didier Cayrac**
Operations Support & Mission Optimization Dept.
Matra Marconi Space, 31, rue des Cosmonautes
31077 Toulouse Cedex – France
E-mail: cayrac@soleil.matra-espace.fr

**Didier Dubois**  **Henri Prade**
I.R.I.T.–C.N.R.S.
Université Paul Sabatier – Bât. 1R3, 118 route de Narbonne
31062 Toulouse Cedex – France
E-mail: {dubois, prade}@irit.irit.fr

## Abstract

An approach to fault isolation that exploits vastly incomplete models is presented. It relies on separate descriptions of each component behavior, together with the links between them, which enables a focusing of the reasoning to the relevant part of the system. As normal observations do not need explanation, the behavior of the components is limited to anomaly propagation. Diagnostic solutions are disorders (fault modes or abnormal signatures) that are consistent with the observations, as well as abductive explanations.
An ordinal representation of uncertainty based on possibility theory provides a simple exception-tolerant description of the component behaviors. We can for instance distinguish effects that are more or less certainly present (or absent) and effects that are more or less possibly present (or absent) when a given anomaly is present. A realistic example illustrates the benefits of this approach.

## 1 INTRODUCTION AND RATIONALE

Developing models of systems that are both expressive enough to support effective diagnosis and cheap enough to keep knowledge acquisition cost affordable in real world contexts remains an open problem. This seems to prevent a wide acceptance of model-based diagnosis in industry. Models of the system to be diagnosed traditionally consist of a description of its structure, in terms of components linked together, and of a description of the behavior of the individual components. On the one hand, acquisition of the structure is "cheap", as it can often be derived directly from design information (the detection of faults altering the structure of the system cannot be handled in this framework). On the other hand, the behavior of the individual components of the system is much more difficult to model in an appropriate way. Two approaches are usually followed: the correct behavior and / or the fault modes of each component are represented. The representation choice is conditioned to a large extent by the diagnostic approach chosen: consistency-based or abductive.

Consistency-based diagnosis (Davis and Hamscher, 1988, deKleer and Williams, 1987) exploit a nominal behavior model to derive explanations that allow the restoration of the consistency of the predicted behavior with the observations. This approach requires an expensive model because it must allow an effective detection of the discrepancies between predicted and observed behaviors (i.e., real anomalies are indeed detected, and there are few spurious ones). We believe that the cost of a model capable of such a high quality simulation (and expressed in first order logic) would be prohibitive for most real world applications (except maybe digital circuitry), if it were even feasible.

Abductive diagnosis (Poole, 1989) relies on a causal model between faults and manifestations, and looks for explanations (faults) that cause, or "cover" the symptoms observed. Explanation as consistency is a weaker notion than explanation as covering: the consistent explanations form a superset of the abductive explanations. Abductive diagnosis requires completeness of the identified fault modes (Console and Torasso, 1990). Proposals for handling incompleteness (Console et al., 1989) still require the identification of all possible sources of incompleteness, at the level of each rule, which makes their application unfit to vastly incomplete models.

If all observations were to be explained abductively, a model of the nominal *and* fault behavior of the system that is just as fine as for the consistency-based approach would be required. However, Console and Torasso also show that if only part of the observations (for instance, the abnormal ones) need to be abductively explained, a model describing how anomalies are propagated is sufficient.

Section 2 distinguishes between fault detection and fault isolation, which require different kinds of knowledge. The approach to fault isolation we outline exploits vastly incomplete models, and allows the identification of fault modes which are not explicitly described. An ordinal representation of uncertainty based on possibility theory allows a qualitative and incomplete description of the component behaviors, at an appropriate level of abstraction. Section 3 presents the possibilistic logic treatment. The proposed approach is formalized in Section 4, and illustrated by a realistic example in Section 5.

## 2 OVERVIEW OF THE APPROACH

The diagnostic problem at large (which may be summarized by "what is wrong with my system ?") includes two goals: determining what is wrong from the "outside" point of view, i.e., identifying the symptoms, and determining what is wrong from the "inside" point of view, i.e., identifying the part of the system that is originally responsible for the symptoms. These two tasks (which often are not distinguished) actually make best use of different kinds of knowledge, and thus would benefit from being treated separately, with distinct models of the system. Indeed, sets of simple nominal behavior models of restricted scope are often sufficient for monitoring; besides, simple influence models describing anomaly propagation are sufficient for fault isolation if a well-focused inference is used.

## 2.1 FAULT DETECTION

In order to detect a fault (i.e., at first, to detect that something is wrong with the system), a model of the nominal, i.e., correct, behavior of the system is necessary. To put it plainly, *"you can't know that something is wrong if you don't know what the system is supposed to do when it is ok"*. This task is often performed by some dedicated monitoring facilities. In the case of satellites, the telemetry flow is monitored in real time. Typically, the value of some observable parameter is compared to some thresholds, or its evolution is matched to a standard pattern. This model of the nominal behavior is thus a lot simpler than a simulator (either qualitative or numerical), and therefore preferred, as it is sufficient to fulfill the monitoring task.

The output from the monitoring phase is a first set of symptoms (typically abnormal states of variables, or abnormal evolutions). Further analysis performed off line (for instance through additional probing or finer analysis of collected data) allows the discovery of additional symptoms. As this additional analysis is expensive, it should be focused, and driven by the fault isolation task.

This simple approach has only one drawback: using Model-Based terms, it generates discrepancies, but not conflicts (minimal sets of components of which at least one is faulty): fault hypotheses have to be generated in the fault isolation phase.

## 2.2 FAULT ISOLATION

At this stage, given abnormal observations, we are looking for explanations in terms of anomalies inside the system, i.e., fault(s) of some parts of the system. The absence of a full-fledged nominal behavior model rules out the application of the conventional consistency-based approach. Abductive approach only requires a model describing how anomalies are propagated, but, as pointed out, it still demands exhaustive identification of:
- fault modes of components
- influences within components (i.e., how an anomaly in input of a component affects its outputs).

These two constraints are relaxed, allowing more incompleteness in the model, while recovering all relevant explanations through focused consistency-based and/or abductive reasoning on the subset of the model relevant to the symptoms.

### 2.2.1 IDENTIFICATION OF "NEW" FAULT MODES

In addition to conventional "abducibles" formed by identified fault modes of each component (e.g., relay "stuck at 1", or "stuck at 0"), we also accept as diagnostic solutions abnormal states of the outputs of a given component. For instance, "noisy output" of a filter may be accepted as a fault hypothesis, even if no identified fault mode specifically predicts this abnormal output. This allows the characterization of fault modes that were not identified. The "new" fault modes are characterized by an abnormal signature on the component outputs. This information may be sufficient for repair, or an expert may decide whether they correspond to possible fault modes.

Pre-identified fault modes may be preferred, and thus get a higher priority in the discrimination. Among fault hypotheses formed by abnormal output of components, those associated to components for which the anomaly cannot be explained by anomalies located further upstream may be preferred. This heuristic preference order allows a better exploitation of the results (e.g., "conventional" abducibles come first).

The possibility of accepting as candidates the characterisation of possible abnormal behavior improves the exploitation of incomplete models, while the preference heuristics allow a sorting of the potentially numerous hypotheses generated.

### 2.2.2 INCOMPLETE MODEL EXPLOITATION

To relax the completeness assumption on influences (anomaly propagation) within components, we propose to use a form of consistency-based reasoning on a *relevant* subset of the model, in complement to abductive reasoning. This relevant subset corresponds to a restriction of the model to the parts which are somehow related to observed symptoms (a formal definition of this approach is given in Section 4). A *relevant consistent explanation* is then a fault mode or an abnormal signature that belongs to this subset, and is consistent with all the observations.

Introducing uncertainty in the behavior representation avoids the need for complete elicitation of exceptional cases. It is also in agreement with incompleteness of the description of the diagnosed situation. In the following, we propose an ordinal representation of uncertainty of the behavior of individual components (Dubois and Prade, 1993, Cayrac et al., 1994a,b). It allows the expression of:
- (more or less) certain influences between inputs and outputs (possibly taking into account the configuration of the component, e.g., "on" or "off" for a relay)
- (more or less) impossible influences (e.g., a given input cannot lead to a given output)

It thus leaves room for influences that are (more or less) possible.

The basic principle of the reasoning used at the local (component) level is as follows: all abnormal inputs that *may* cause the abnormal output (i.e., the influence is not impossible) are gathered: they are relevant consistent explanations. Among these, the ones that (more or less) certainly cause the anomaly (i.e., the influence is certain) are preferred: they are abductive explanations.

### 2.2.3 THE FAULT ISOLATION PROCESS

- *Hypotheses generation*
  As no hypothesis is provided by the monitoring phase, it is necessary to generate them as a first step of fault isolation. We want to find disorders that are directly or indirectly "related" to the observed manifestations, i.e., components which may be responsible for the symptoms. The fact that a disorder is "related" to a manifestation can have two meanings: a strong (abductive) meaning, in which the hypotheses entails the manifestation, or a weaker (consistency-based) meaning, in which the hypothesis is just consistent with the symptom and there is a possible influence path between them (this notion will be formalized in the next section). We thus have to follow upstream "influence paths", and collect as candidates the acceptable disorders found on the way.

- *Additional manifestations prediction:*
  The monitoring phase only provides an incomplete set of symptoms. It is therefore useful to generate additional expected manifestations for the fault hypotheses, in order to allow discrimination.

- *Probing and hypotheses discrimination:*
  The expected manifestations are tested, and whether these new observations are entailed by or are consistent with the disorders is verified. This allows the update of the plausibility of the disorders: an inconsistent hypothesis will be rejected, a covering hypothesis (abductive solution) will be preferred over hypotheses that are just consistent with the symptoms.

This discussion suggests that a simple fault influence propagation model may be sufficient for the fault isolation task, given proper focusing. The ideas introduced are formalized in the next section.

## 3  POSSIBILISTIC LOGIC APPROACH

In possibilistic logic (Dubois et al., 1994a,b), classical logic formulas are weighted by lower bounds of the necessity degree with which the formula is held for true. A weighted formula $(\varphi,\alpha)$ is thus interpreted as the semantic constraint $N(\varphi) \geq \alpha$ where N is a necessity measure. A necessity measure is associated with a plausibility ordering of the possible interpretations of the world encoded by a so-called [0,1]-valued possibility distribution $\pi$. $\pi(\omega) > \pi(\omega')$ means that interpretation $\omega$ is more plausible than interpolation $\omega'$. It is assumed that $\exists \omega, \pi(\omega)=1$. Then $N(\varphi)=\min\{1-\pi(\omega)|\omega\models \neg\varphi\}=1-\max\{\pi(\omega) | \omega\models \neg\varphi\}=1-\Pi(\neg\varphi)$ where $\Pi$ is the possibility measure (Zadeh, 1978) associated with $\pi$ and the necessity of $\varphi$ corresponds to the impossibility of $\neg\varphi$. It follows that $N(\varphi \wedge \psi) = \min(N(\varphi),N(\psi))$. $N(\varphi) = 1$ indicates that it is certain that $\varphi$ is true $N(\varphi) = 0$ indicates that $\varphi$ is unknown.

Automated reasoning can be performed by means of an extended resolution principle in possibilistic logic:

$$N(\varphi) \geq \alpha, N(\psi) \geq \beta \vdash N(\text{Resolvent}(\varphi,\psi)) \geq \min(\alpha,\beta).$$

A model-based diagnosis problem can be characterized by a logical theory SD describing the behavior of the system, a set of atoms CXT describing contextual data expressing the configurations of the components, and a set of literals OBS representing the observations to be explained. Two kinds of solutions are then defined, with $\mathcal{T}=\text{SD}\cup\text{CXT}$:

- consistency-based approach: an explanation H (this definition will not be elaborated here) is a solution iff:
$$\mathcal{T}\cup\text{OBS}\cup\text{H is consistent}$$

- abductive approach: an explanation H is a solution iff
$$\mathcal{T}\cup\text{H} \vdash \text{OBS}$$

The extension of the two above approaches to the handling of uncertainty using possibilistic logic is now briefly outlined.

Let us consider a knowledge base $\mathcal{T}$ made of uncertain implication relations between possible disorders (explanations) $d_i$, i=1,m and effects, i.e., present or absent, manifestations $m_j$, j=1,n. This is encoded by weighted clauses of the form

$$N(\neg d_i \vee m_j) \geq \alpha_{ij}>0 \text{ and } N(\neg d_k \vee \neg m_r) \geq \lambda_{kr}>0.$$

Besides, we have a set OBS of uncertain observations represented by weighted formulas of the form

$$N(m_s) \geq \beta_s > 0 \quad \text{(present manifestations)}$$
$$N(\neg m_t) \geq \rho_t > 0 \quad \text{(absent manifestations)}.$$

It is assumed that $\forall i, \forall j, \forall r, N(\neg d_i \vee m_j)>0$ and $N(\neg d_i \vee \neg m_r)>0$ entail $j \neq r$, i.e., the behavioural knowledge is coherent. In the approach presented in this paper, $\mathcal{T}$ can be interpreted as the behavior of a given component, the disorders being fault modes and abnormal inputs, and the manifestations being abnormal outputs. Applying the possibilistic resolution rule, we obtain

$$\forall j, N(\neg d_i) \geq \min(\alpha_{ij},\rho_j) \text{ and } \forall r, N(\neg d_i)\geq\min(\lambda_{ir},\beta_r)$$

and thus

$$N(\neg d_i) \geq \max[\max_j \min(\alpha_{ij},\rho_j), \max_r \min(\lambda_{ir},\beta_r)] \quad (1)$$

Thus from (1) we compute an upper bound of possibility degree of the explanation $d_i$. $\Pi(d_i)=1-N(\neg d_i)$ which is the *level of consistency* $\text{cons}_\mathcal{T}(d_i;\text{OBS})\geq 1-N(\neg d_i)$ of $\mathcal{T}\cup\text{OBS}\cup\{d_i\}$ in the possibilistic setting. Indeed, in possibilistic logic the level of inconsistency (the complement to 1 of the level of consistency) of a knowledge base $\mathcal{K}\cup\{\varphi\}$ is nothing but the greatest lower bound $\alpha$ such that $N(\neg\varphi)\geq\alpha$ is compatible with the constraints on the necessity measure N encoding the pieces of knowledge in $\mathcal{K}$.

Besides, the possibilistic logic resolution rule applied to $N(\neg d_i \vee m_j)\geq\alpha_{ij}$ and $N(d_i)=1$ (assuming that $d_i$ is present) entails $N(m_j) \geq \alpha_{ij}$. Let $M^+$ be the set of present manifestations such that $N(m_j)\geq\beta_j>0$. Then in *abductive* reasoning we are interested in finding the $d_i$'s such that $\forall m_j \in M^+$, $\mathcal{T} \cup \{d_i\} \vdash m_j$. When the $\alpha_{ij}$'s and $\beta_j$'s are 0 or 1, $d_i$ is an abductive explanation of $m_j$ if and only if

$$\beta_j \rightarrow \alpha_{ij} = 1 \qquad (2)$$

where $a \rightarrow b = 1$ if $a \leq b$ and $a \rightarrow b = 0$ if $a > b$. This holds for any $m_j$ in $M^+$, and $d_i$ is an abductive explanation of $M^+$ if and only if

$$\Delta^*(d_i) = \min_j (\beta_j \rightarrow \alpha_{ij}) \qquad (3)$$

is equal to 1 (otherwise it is zero). When $\alpha_{ij}$ and $\beta_j$ take values intermediary between 0 and 1, (3) can be extended using Gödel implication $a \rightarrow b = 1$ if $a \leq b$ and $a \rightarrow b = b$ if $a > b$. Then (3) expresses a coverage of the fuzzy set of manifestations more or less certainly observed by the fuzzy set of manifestations more or less certainly caused by $d_i$. Note that $\Delta^*$ is not a necessity degree strictly speaking since $\Delta^*(d_i)=1$ means only that $d_i$ is highly plausible. Note that $\Delta^*(d_i) = 1$ as soon as $\beta_j \leq \alpha_{ij}$, $\forall j$, i.e., when $N(m_j) \geq \beta_j$, is a set of valid deductions from the weighted clauses $N(\neg d_i \vee m_j) \geq \alpha_{ij}$ and $N(d_i)=1$.

In the computation of abductive solutions, only observed manifestations are accounted for. Indeed, we do not wish to explain the nominal behavior. Moreover, negative causal knowledge is often scarce (i.e., many $\lambda_{ir}$ will be 0 in practice). It is easy to see that when $\lambda_{ir}=0$, accounting for an absent manifestation ($\rho_r > 0$) will destroy the abductive information in (3) since then $\rho_r \rightarrow \lambda_{ir}$ is equal to 0.

## 4 FORMAL DESCRIPTION

The model of the system consists of a description of its individual components and their behavior, and of the links defining the relation between them.

### 4.1 COMPONENT DESCRIPTION

The components have identified inputs, outputs, configuration modes (i.e., states), and possibly identified fault modes. All these can be viewed as the parameters of the model. They will be represented as predicates defined on discrete domains.

**Def.1:** A **component** $C_i$ is characterized by:
- A set of predicates describing:
  - its input $\{c_i\_in_1,...,c_i\_in_m\}$,
  - its output $\{c_i\_out_1,...,c_i\_out_p\}$,
  - its configuration mode $c_i\_state$ (if applicable),
  - its fault mode $c_i\_fault$.
- A theory $CD_i$ concerning these predicates, expressed in possibilistic logic and describing its behavior:
  - how an abnormal input signature affects its outputs, given its configuration mode,
  - the impact of each identified fault mode on its outputs

Component is used here as a generic term: a component may be an actual electric component, or a whole equiment, or a function, etc.

Input and output are defined on a set of abnormal states such as "absent", "noisy", "at_zero", etc. Nominal states need not to be represented, as they are not propagated. We consider that the system is static, i.e., that the configuration mode of the component does not change during the diagnostic session. Example of configuration modes of a relay: on/off. Examples of fault modes: stuck_at_zero, stuck_at_one, always_on, etc.

As we only require the explanation of abnormal observations, the behavior representation proposed (anomaly propagation) is sufficient. The information regarding the behavior of each component is kept separate, as this "partitioning" is exploited by the diagnostic task.

It should be noted that even if no gradual representation of uncertainty was used, we could distinguish between influences that are certain and those which are only possible. E.g., in Figure 1, the behavior of $C_1$, "$i_1$ cannot cause an anomaly on $o_1$, but may impact $o_2$, and will impact $o_3$" would be expressed as $CD_1 = \{i_1(abnormal) \rightarrow \neg o_1(abnormal), i_1(abnormal) \rightarrow o_3(abnormal)\}$. At the component level, $i_1(abnormal)$ is an abductive explanation of $o_3(abnormal)$, a consistent explanation of $o_2(abnormal)$, and there is no explanation for $o_1(abnormal)$. Note: in this simple example, the component has only one input; in the general case, an abnormal input signature should be considered.

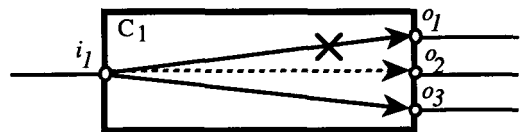

Figure 1

The use of both consistency-based and abductive approaches at the component level captures of a form of incompleteness. However, as mentioned previously, a greater expressivity of uncertainty is desirable. We may for instance want to express that a fault mode is more likely to impact one output of the component than another, or distinguish unlikely impacts from those which are plainly impossible. The goal of this refinement is to allow a ranking of the diagnostic solutions, for instance through the identifications of fault hypotheses that are more or less consistent with the observations, or that more or less cover them.

The behavior $CD_i$ of the component $C_i$ can thus be equivalently defined by:
- a set of uncertain clauses encoded in possibilistic logic,
- a global relation on the Cartesian product of the set of variables formed by the input, output, configuration mode and fault mode of the component, weigthed in terms of necessity degrees (Cayrac et al. 1994a).

A subset of the outputs of the components is assumed to be observable. However, some observations may be pervaded with uncertainty.

## 4.2 LINKS BETWEEN COMPONENTS

The links describe the way in which the components are connected, and characterize their possible interactions. As components have distinct inputs and outputs, the links are directed. They just carry the value of the output of a component to some inputs of other components, without modification.

**Def.2:** A **link** is a proposition of the form:
$c_i\_out_j(x) \rightarrow c_k\_in_l(x) \wedge \ldots \wedge c_p\_in_q(x)$
expressing that the output state $c_i\_out_j(x)$ of the component $C_i$ is propagated to the inputs $c_k\_in_l(x), \ldots, c_p\_in_q(x)$ of components $C_k, \ldots, C_p$.

**Def.3:** The set **LINKS** contains all the links between the components of the system.

Although this is usually not done in most model-based diagnostic approaches, we choose to identify the links as individual entities. Indeed, in case several components are related through more that one possible interaction path (e.g., two links exist between them), this allows to focus the reasoning to the relevant links.

As defined, the links only allow distribution of information (fan-out), but no fusion (fan-in). This limitation allows a clear separation between components, which encapsulate all the behavior, and the "wiring" represented by links. In a fan-in situation, when the effects can be superposed (they do not interfere) the relation can be represented by several links; when they do interfere, an additional component must be created.

## 4.3 THE DIAGNOSTIC PROBLEM

**Def.4:** The **model of the system, SD,** is formed by the component behaviors and the links between them:
$SD = \{CD_1, \ldots, CD_n, LINKS\}$

**Def.5:** We call **context, CXT,** the set of configuration modes of the components at the time of diagnosis. This is assumed to be known a priori: $CXT = \{c_i\_state(state_k), \ldots\}$

**Def.6:** We call **set of observations, OBS,** the set of the output states which have been observed. OBS is partitioned into two subsets $M^+$ (manifestations whose presence is confirmed) and $M^-$ (those whose absence is confirmed).
$OBS = \{c_i\_out_h(state_q), \ldots, \neg c_j\_out_k(state_t), \ldots\}$
$\quad = M^+ \cup M^- = \{m_1, \ldots, \neg m_n, \ldots\}$

$c_5\_out_1$(noisy) is an example of a present manifestation. It should be noted that an observation is here equivalent to an abnormal state of a link, as a link is tied to exactly one component output. The observations may also be more or less certain. $M^+$ and $M^-$ are in fact fuzzy sets whose membership degrees are interpreted as certainty levels: $N(m_1) \geq \mu_{M^+}(m_1), \ldots, N(\neg m_n) \geq \mu_{M^-}(m_n), \ldots$

**Def.7:** A **diagnostic problem** is the tuple **DP**, formed by the model of the system, its configuration at the time of diagnosis, and the set of observations:

$DP = \{SD, CXT, OBS\}$.

In addition to a possibilistic version of "conventional" abductive reasoning, we propose a form of possibilistic consistency-based reasoning. It exploits a theory formed by the behaviors of components that are on an upstream influence path and the relevant links between them. The following two definitions characterize its elements.

The **set of possible influence paths leading to a manifestation m** (given output state $c_p\_out_q(state_r)$ of a component $C_p$) is the subset **REL_LINKS(m)** of LINKS containing links that may propagate the cause of m. REL_LINKS(m) is defined recursively as follows:
a link $c_i\_out_j(x) \rightarrow c_k\_in_v(x) \wedge \ldots \wedge c_s\_in_t(x)$ of LINKS belongs to REL_LINKS if and only if:

– it is tied to an input $c_p\_in_\tau(x)$ of $C_p$, and there is at least one state of this input that is (at least partially) consistent with m, i.e., with $c_p\_out_q(state_r)$:

  $p \in \{k, \ldots, s\}$, and $\exists \tau, \exists state_\gamma, CD_p \cup \{c_p\_in_\tau(state_\gamma)\}$ is consistent to a strictly positive degree.

– it is tied to the input $c_\omega\_in_\tau(x)$ of a component $C_\omega$ of which an output $c_\omega\_out_\lambda$ matches the premise of one of the links of REL_LINKS, and there is at least one state $s$ of $c_\omega\_in_\tau$ that is at least partially consistent with m and the model of the system (i.e., there is a possible influence path between $c_\omega\_in_\tau$ and $c_p\_out_q(state_r)$):

  $\exists \omega \in \{k, \ldots, s\}, \exists \lambda, \exists state_\gamma,$
  $c_\omega\_out_\lambda(x) \rightarrow \ldots \in $ REL_LINKS(m) and
  $SD \cup \{c_\omega\_in_\tau(state_\gamma)\} \cup \{c_p\_out_q(state_r)\}$ is consistent to a stictly positive degree.

REL_LINKS (relevant links) forms a restriction of the theory LINKS to the links that are relevant to a given symptom. This restriction is performed on the basis of possible influence paths *between* components (links), and *inside* them.

The set **REL_COMPS(m) of components relevant to a manifestation m** (given output state $c_p\_out_q(state_r)$ of $C_p$), contains $C_p$ and all the components of which an output is tied to a possible influence path to m:
REL_COMPS(m) = $\{C_i,$ i=p or
  $\exists \lambda, c_i\_out_\lambda(x) \rightarrow \ldots \in $ REL_LINKS(m)$\}$

REL_COMPS(m) contains all the components that may be responsible for the state observed on the output of a given component (manifestation m).

## 4.4 SINGLE FAULT SOLUTIONS

An **elementary solution d of the diagnostic problem** DP is either:
• a fault mode d=$c_i\_fault(fault\_mode_j)$ of component $C_i$, or
• a set of abnormal states d=$\{c_i\_out_j(state_k), \ldots\}$ on the outputs of $C_i$, (and d$\cap$OBS=$\emptyset$: d is not a trivial solution), such that:
• d causes more or less certainly all present manifestations:

$SD \cup CXT \cup \{d\} \vdash M^+$ with a strictly positive necessity degree $\alpha$, and $SD \cup CXT \cup \{d\} \cup M^-$ is completely consistent with the observations d is then an **abductive explanation to the degree** $\alpha$ of the abnormal observations.

- there is a possible influence path between d and each present manifestation, and d is not completely inconsistent with the observations:

$d=c_i\_fault(fault\_mode_j)$; $\forall m \in M^+$, $C_i \in REL\_COMPS(m)$ and $SD \cup CXT \cup \{d\} \cup OBS$ is consistent to a srictly positive degree $\beta$

or:

$\forall \lambda$, $\forall s$ such that $c_i\_out_\lambda(s) \in d$, $\forall m \in M^+$,

$c_i\_out_\lambda(x) \to \ldots \in REL\_LINKS(m)$, and

$SD \cup CXT \cup \{d\} \cup OBS$ is consistent to a strictly positive degree $\beta$.

d is then a **relevant explanation consistent to the degree** $\beta$ with the observations.

Clearly, in the above expressions, SD can be replaced by its relevant (useful) parts:

$\cup_{m \in M^+}[REL\_LINKS(m) \cup REL\_COMPS(m)]$.

The single fault solutions are thus fault modes or abnormal output signatures which:

- *(more or less)* entail all the manifestations (abductive explanations)
- are *(more or less)* consistent with the observations, and there exists an influence path between the hypothesis and each manifestation.

Thus, a ranking of more or less plausible solutions to the diagnosis problem is obtained.

### 4.5 PROBING POINTS

An **expected manifestation** m with respect to a solution d of a diagnostic problem DP, any abnormal observable output state of a component, $m=c_i\_out_j(state_k)$, such that:

$SD \cup CXT \cup \{d\} \vdash m$ with a strictly positive necessity degree.

m is then expected to be present when d is present.

or:

$SD \cup CXT \cup \{d\} \vdash \neg m$ with a strictly positive necessity degree.

m is then expected to be absent when d is present.

The observation of whether an expected manifestation m is present or not allows the diagnostic system to check whether d is still a consistent and / or an abductive explanation: m is added to OBS.

## 5 A REALISTIC EXAMPLE

### 5.1 A SATELLITE SOLAR ARRAY EQUIPMENT

The goal of the power regulation subsystem is to meet the power requirements of the equipments connected on the bus. A ladder of comparators connects as many solar arrays (S.A.) as necessary (Figure 2). When the voltage is too high on the power bus, S.A.'s are progressively disconnected. If, due to an anomaly, this is not enough, a protection resistor is connected to create an extra load. On the contrary, if all S.A.'s are connected and the power demand still cannot be met, the control system assumes that the sun is no longer visible (it is behind the Earth), and sends an eclipse signal to an on-board computer, so that batteries can be used to supply extra power. A simplified view of the contol system is given in Figure 3.

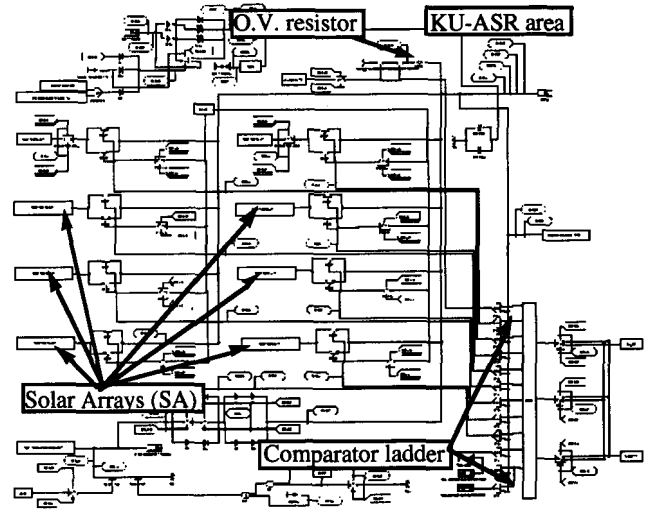

Figure 2

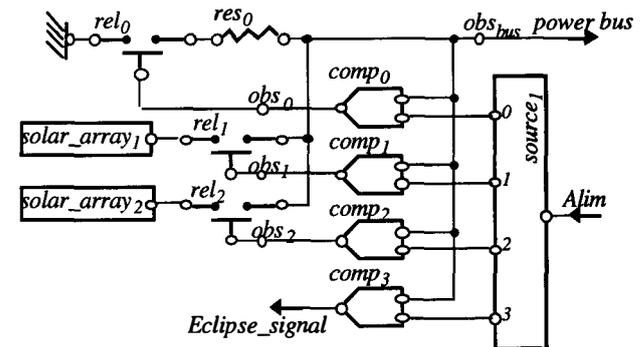

Figure 3

### 5.2 MAIN COMPONENTS

In this example, we distinguish between analog and digital input/output. Analog i/o identified states are *DEG* and *ABS* (respectively for degraded and absent; only abnormal states are used). Digital i/o identified states are *ZERO*, *ONE*. Similar extremely simple sets of qualification turned out to be a good tradeoff between knowledge acquisition cost and solutions quality for the modeling of complex systems.

We choose to represent explicitly (more or less) certain and (more or less) impossible influences between inputs and outputs of components. The uncertainty is limited to the following qualitative levels: "certain", "almost certain", "likely; "possible" (i.e., unknown); "unlikely", "almost impossible", "impossible". In other words, we use a discrete linearly ordered scale. In practice, levels of certainty of presence of, for instance, a manifestation m ($\mu_{M^+}(m)$), can be represented by a number: 1.0 for "certain", 0.8 for "almost certain",..., 0.0 for "possible"; the levels of certainty of absence of m can be similarly encoded: 1.0 for "impossible", 0.8 for "almost impossible",..., 0.0 for "possible". Note that only the ordering between these numbers is meaningful, and not their exact value. Since only purely ordinal operations such as min, max, order-reversing and Gödel implication are used, the formulas used can be straightforwardly transposed from the [0,1] scale to a discrete scale.

Behavior of the components used in the simplified example:

**Comparator**: $comp_i$

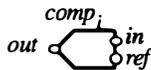

input: ref (analog), in; (analog) output: out. (digital)
behavior:

$comp_i\_ref(ABS) \rightarrow comp_i\_out(ONE)$, certain

$comp_i\_ref(ABS) \rightarrow comp_i\_out(ZERO)$, impossible

$comp_i\_ref(DEG) \rightarrow comp_i\_out(ONE)$, likely

$comp_i\_in(ABS) \rightarrow comp_i\_out(ZERO)$, certain

$comp_i\_in(ABS) \rightarrow comp_i\_out(ONE)$, impossible

*Note: the behavior of the comparator is fully described when its "ref" or "in" input is ABS. However, when it "ref" input is DEG, the output may be ONE (likely) or, implicitly, ZERO (possible); when its "in" input is DEG, any output state is possible.*

**Solar array section**: $solar\_array_i$

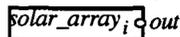

output: out. (analog)

**Relay**: $rel_i$

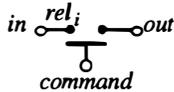

input: command (digital), in; (analog) output: out. (digital)
behavior:

$rel_i\_state(ON) \wedge rel_i\_in(ABS) \rightarrow rel_i\_out(ABS)$, certain

$rel_i\_state(ON) \wedge rel_i\_in(DEG) \rightarrow rel_i\_out(DEG)$, certain

$rel_i\_state(OFF) \wedge rel_i\_in(ABS) \rightarrow rel_i\_out(ABS)$, impossible

$rel_i\_state(OFF) \wedge rel_i\_in(DEG) \rightarrow rel_i\_out(DEG)$, impossible

*Note: nothing is said about the response of the relay to an ON or OFF command input since we do not represent the dynamic behavior of the components.*

**Voltage source**: $source_i$

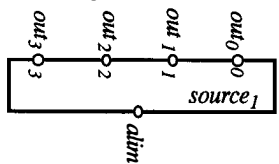

*The voltage source gives in output a scale of voltages used as references for comparators. Its input is its alimentation.*
input: alim (analog) output: $out_1$, $out_2$, $out_3$, $out_4$ (analog)
behavior:

for all j, $source_i\_alim(ABS) \rightarrow source_i\_out_j(ABS)$, certain

$source_i\_alim(DEG) \rightarrow source_i\_out_0(DEG)$, unlikely

$source_i\_alim(DEG) \rightarrow source_i\_out_2(DEG)$, likely

$source_i\_alim(DEG) \rightarrow source_i\_out_3(DEG)$, likely

**Ground**: $ground_i$ 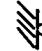

output: out. (analog)

**Resistor**: $res_i$ 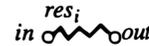

input: in; (analog) output: out. (analog)
behavior:

$res_i\_in(DEG) \rightarrow res_i\_out(DEG)$, almost certain

*Note: nothing is said about the impact of an ABS input: the output may be ABS or DEG or neither one (all alternatives are completely possible)*

### 5.3 DIAGNOSTIC SESSION

Initial symptom: $OBS=M^+=\{(Eclipse\_signal(ONE),$ certain)}: observation of an eclipse signal, which is abnormal because at this time the satellite is still facing the sun. $CXT=\{rel_0(OFF), rel_1(ON), rel_2(OFF)\}$

*Hypotheses generation phase*

Figure 4 shows the theory relevant to the original symptom. Hypotheses (which in this example consist only in abnormal output of some components) are denoted in boldface.

Abductive explanations, computed using the Formula (3):
$D^*=\{(source\_out_3(ABS),1.0), (alim(ABS),1.0),$
$(source\_out_3(DEG),0.3), (alim(ABS),0.3)\}$

$Eclipse\_signal(ONE)$ is a certain consequence of $source\_out_3(ABS)$ and $alim(ABS)$, but only a likely consequence of $source\_out_3(DEG)$ and $alim(DEG)$. The former explanations, which are abductive explanations in the conventional sense, are thus preferred at this point.

Consistent explanations, computed using the Formula (1) (all of them are fully consistent with the eclipse signal symptom):

$\tilde{D}=\{(source\_out_3(ABSENT),\text{possible}),$
$(source\_out_3(DEGRADED),\text{possible}),$
$(res_0\_out(DEGRADED),\text{possible}),$
$(rel_1\_out(DEGRADED),\text{possible}),$
$(rel_2\_out(DEGRADED),\text{possible}),$
$(rel_0\_out(DEGRADED),\text{possible}),$
$(rel_0\_out(ABSENT),\text{possible}),$
$(solar\_array_1\_out(DEGRADED),\text{possible}),$
$(solar\_array_1\_out(ABSENT),\text{possible}),$
$(alim(ABSENT),\text{possible}),$
$(alim(DEGRADED),\text{possible})\}$

*Additional manifestation prediction phase*

A prediction phase allows the elicitation of additional manifestations that may further discriminate among pending hypotheses. Figure 5 summarizes this phase of the reasoning.

The additional manifestations predicted by the various fault hypotheses are therefore: $obs\_bus(DEG)$, $obs\_bus(ABS)$,

$obs_0(ONE)$, $obs_1(ONE)$, $obs_2(ONE)$.

*Additional manifestation probing phase*

We observe: {($obs\_bus(DEG)$, almost certain), ($obs\_bus(ABS)$, impossible), ($obs_0(ONE)$, impossible), ($obs_1(ONE)$, certain), ($obs_2(ZERO)$, certain)}.

($obs_0(ZERO)$, certain), which is equivalent to ($obs_0(ONE)$, impossible) is completely inconsistent with $alim(ABS)$, which predicts ($obs_0(ONE)$, certain) and ($obs_0(ZERO)$, impossible), and partially inconsistent with $alim(DEG)$, which predicts ($obs_0(ONE)$, likely). The hypothesis $alim(ABS)$ is therefore discarded and $alim(DEG)$ is unlikely. Similarly, ($obs\_bus(ABS)$, impossible) is partially inconsistent with $solar\_array_1\_out(ABSENT)$, which is therefore considered unlikely.

The other hypotheses remain completely possible, i.e., they are located on a possible influence path leading to some of the symptoms, and they are completely consistent with all the observed symptoms. However, no abductive hypothesis remains: $source_1\_out_3(ABS)$ and $source\_out_3(DEG)$ do not abductively explain the symptom ($obs\_bus(DEG)$, almost certain).

*Conclusion of the diagnosis session - exploitation of the results*

Abductive explanations: $D^* = \emptyset$
Consistent relevant explanations:

$\widehat{D} = \{(source\_out_3(ABS), \text{possible}),$
$(source\_out_3(DEG), \text{possible}),$
$(rel_0\_out(ABS), \text{possible}), (rel_0\_out(DEG), \text{possible}),$
$(res_0\_out(DEG), \text{possible}),$
$(solar\_array_1\_out(DEG), \text{possible}),$
$(rel_1\_out(DEG), \text{possible}), (rel_2\_out(DEG), \text{possible}),$
$(solar\_array_1\_out(ABS), \text{unlikely}),$
$(alim(DEG), \text{unlikely})\}$

$source\_out_3(ABS)$, $source\_out_3(DEG)$ are rejected by the user because they cannot explain the abnormal bus voltage. $solar\_array_1\_out(DEG)$ is relevant: if the array is damaged, it will produce less power, and this may trigger the eclipse signal. It is preferred over $rel_1\_out(DEG)$, which can be one of its manifestations. $rel_2\_out(DEG)$ is quickly discarded by the user: an extra solar array connected cannot produce the symptoms observed. $rel_1\_out(DEG)$ is actually irrelevant, since the relay is off.

$rel_0\_out(ABS)$ and $rel_0\_out(DEG)$ are preferred over ($res_0\_out(DEG)$, possible), which may be their manifestations. $rel_0\_out(ABS)$ is in fact the explanation that is the closest to the real fault. Indeed, $rel_0$ is actually faulty: it connected unduly $res_0$ to the power bus, causing the voltage to go down, and, as the bus was heavily loaded, triggering the eclipse signal.

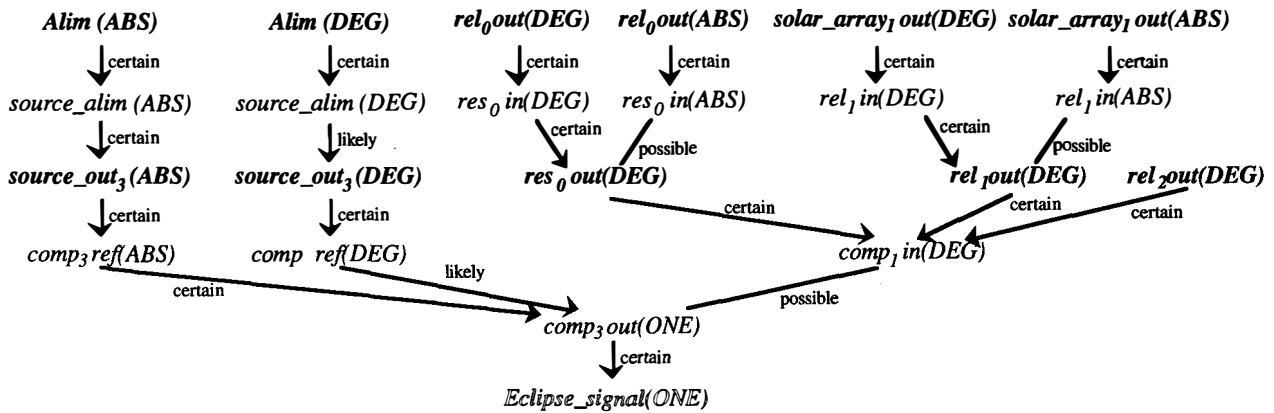

Figure 4 Subset of the theory relevant to the symptom Eclipse_signal(ONE)

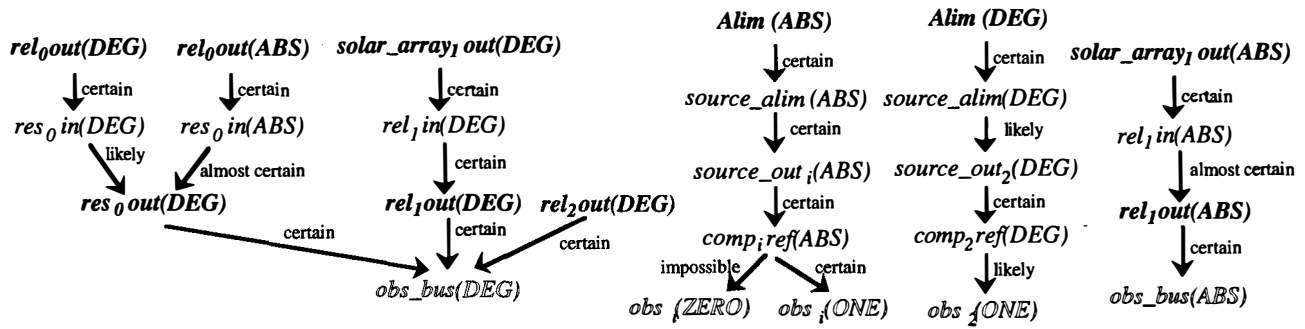

Figure 5 Additional manifestation prediction phase

## 6 CONCLUSION

Preliminary ideas which led to the proposed approach were introduced in (Haziza, 1988), and experimented through the DIAMS project led by Matra Marconi Space since 1987. Its goal is the delivery of operational diagnostic support systems for satellites. Early experiments of model-based approaches were judged not fully satisfactory, especially with respect to the knowledge acquisition costs. During the last eight years, several systems were developed, based on the joint use of various diagnostic knowledge: fault trees, causal knowledge and functional (model-based) knowledge. They include one pre-operational version installed in the Telecom-2 satellite control center early 1994 (Brenot et al., 1993).

In this paper, we proposed a route that may improve the feasibility of operational applications of model-based reasoning. It allows the exploitation of a vastly incomplete model through well focused diagnostic strategies. However, it clearly falls in the scope of diagnostic *support* systems, as the number of possible hypotheses generated is clearly larger than if a complete model were available. However, we recover part of the lost discrimination power through the introduction of an ordinal treatment of uncertainty which enables to rank-order the remaining candidates. Possibilistic logic seems to be well-adapted to contexts in which information is scarce or expensive to gather (significant incompleteness, absence of priors).